\documentclass[runningheads]{llncs}
\usepackage[T1]{fontenc}
\usepackage{graphicx}
\usepackage{tabularray}
\usepackage{tcolorbox}
\usepackage{wrapfig}
\usepackage{subcaption}
\usepackage{amsmath}

\begin{document}
\title{How Social is It? A Benchmark for LLMs' Capabilities in Multi-user Multi-turn Social Agent Tasks}
%
%

\author{Yusen Wu\inst{1} \and
Junwu Xiong\inst{2} \and
Xiaotie Deng\inst{1}}
\institute{School of Computer Science, Peking University \and
Ant Group Co., Ltd}

%
\maketitle              
\begin{abstract}
Expanding the application of large language models (LLMs) to societal life, instead of primary function only as auxiliary assistants to communicate with only one person at a time, necessitates LLMs' capabilities to independently play roles in multi-user, multi-turn social agent tasks within complex social settings. However, currently the capability has not been systematically measured with available benchmarks. To address this gap, we first introduce an agent task leveling framework grounded in sociological principles. Concurrently, we propose a novel benchmark, How Social Is It (we call it HSII below), designed to assess LLM's social capabilities in comprehensive social agents tasks and benchmark representative models. HSII comprises four stages: format parsing, target selection, target switching conversation, and stable conversation, which collectively evaluate the communication and task completion capabilities of LLMs within realistic social interaction scenarios dataset, HSII-Dataset. The dataset is derived step by step from news dataset. We perform an ablation study by doing clustering to the dataset. Additionally, we investigate the impact of chain of thought (COT) method on enhancing LLMs' social performance. Since COT cost more computation, we further introduce a new statistical metric, COT-complexity, to quantify the efficiency of certain LLMs with COTs for specific social tasks and strike a better trade-off between measurement of correctness and efficiency. Various results of our experiments demonstrate that our benchmark is well-suited for evaluating social skills in LLMs.

\end{abstract}
\section{Introduction}

\subsection{A Subsection Sample}
Large language models (LLMs) enhance their expressive and reasoning capabilities through an increase in model parameters, depth, and breadth. They exhibit robust knowledge retention and reasoning abilities, and are continuously evolving. Recent surveys \cite{zhao2024surveylargelanguagemodels}, \cite{minaee2024largelanguagemodelssurvey}, and \cite{gao2023retrievalaugmented} provide comprehensive and detailed insights into this evolution. In practical applications, LLMs have shown significant potential across various domains, contributing notably to multi-agent systems \cite{han2024llmmultiagentsystemschallenges}, \cite{guo2024large}, \cite{He2024LLMBasedMS}, digital humans \cite{yang2024llmbaseddigitaltwinoptimizing}, \cite{zhang2023advancingzeroshotdigitalhuman}, embodied intelligence \cite{li2024larmlargeautoregressivemodel}, \cite{song2023llmplannerfewshotgroundedplanning}, education, intelligent customer service \cite{xu2024retrievalaugmentedgeneration}, \cite{shi2024chopschatcustomerprofile}, and code generation \cite{jiang2024surveylargelanguagemodels}, \cite{hassid2024larger}, bringing artificial intelligence closer to everyday life.
However, when compared to internet technology, which has become ubiquitous in social interactions through iterative development, there remains a more discernible gap between LLMs developer and non-developer accessibility. For instance, LLMs usually struggle to communicate independently with customers without supervision and do not excel in roles such as daily butler services or managing comprehensive company operations beyond simple tasks. Beyond the underutilization of current computational power, a significant reason may be the lack of capabilities in LLMs to independently and skillfully interact in complex social scenarios. To examine those possibilities we need to do precise evaluation to social capabilities.
To study LLMs' capabilities in complex social tasks is also essential for enhancing LLM sociology analysis. Recent exploration of the rationality and biological traits of LLMs, as discussed in \cite{chen2023emergenceeconomicrationalitygpt} and \cite{lyu2024gpgptlargelanguagemodel}, has been complemented by research simulating virtual societies and systems through dialogues among multiple LLMs. This research aims to analyze their social attributes and perform a social division of labor, as exemplified in works such as \cite{gurcan2024llmaugmentedagentbasedmodellingsocial}, \cite{dai2024artificialleviathanexploringsocial}, and \cite{gao2023s3socialnetworksimulationlarge}. These endeavors, however, may be constrained by their idealized scenario settings, which limit the degree of realism \cite{zhou2024reallifejustfantasy}. By building more complex and closer-to-reality agent tasks in social scenes based on sociology theory, and then benchmarking above them we may get sociology about LLM more solid. 

Up to date the significance of LLMs' interpersonal communication skills has become gradually recognized, but current benchmarks have not fully covered this. To evaluate these skills, some works such as SOTOPIA-EVAL \cite{zhou2024sotopiainteractiveevaluationsocial} and MUCA \cite{mao2024multiuserchatassistantmuca} have been made. SOTOPIA-EVAL focuses on designing scenarios to assess social intelligence through role-playing, comparing models against human performance. MUCA, on the other hand, simulates group interactions to establish a framework for determining chat targets' interactions with specified objects. Both works highlight the importance of multi-user dialogue in social relationships and the need for evaluation. Yet, currently there has been no work to bridge social dialogue scenarios with traditional dyadic dialogue assessments, explore their interrelation from a sociological perspective, and build a systematical benchmark for overall evaluation in all social capability dimensions.Moreover, mainstream evaluation frameworks, including thep \cite{chen2024tevalevaluatingtoolutilization} arena \cite{chiang2024chatbotarenaopenplatform}, GPQA \cite{rein2023gpqagraduatelevelgoogleproofqa}, and security assessment frameworks \cite{zhang2024safetybenchevaluatingsafetylarge} \cite{li2024saladbenchhierarchicalcomprehensivesafety}, also do not explicitly assess social communication capabilities as a distinct dimension alongside mathematical, coding, and other critical thinking skills.

As our first approach to assess social competencies effectively, it is imperative to dissect the foundational elements of social interaction through a sociological lens and reinterpret them within the framework of LLMs. Our approach is bolstered by seminal sociological works that analyze social dynamics \cite{goffman1959presentation}, \cite{duncan1972some}, \cite{clark1991grounding}, alongside contemporary investigations into LLMs from a sociological standpoint \cite{dai2024artificialleviathanexploringsocial},\cite{lan2024llmbasedagentsocietyinvestigation}. Furthermore, the field of artificial intelligence, especially multi-agent systems, has provided valuable insights into the analysis and reconstruction of hierarchical systems \cite{Li2019HierarchicalRL}. By integrating these perspectives, we introduce a tiered division of tasks for social agents, categorizing them into four distinct levels: the first being fundamental and well-explored, the subsequent two being relatively autonomous, and the final level representing an integration of the former two.

With novel LLM sociology framework  We endeavor to develop high-caliber evaluation datasets akin to established benchmarks. Deviating from the common practice of repurposing existing LLM datasets, we opt to initiate our dataset construction with manually and fairly reviewed-filtered real news data. The news data is algorithmically clustered and detoxified to capture a more authentic and representative cross-section of real-world social scenarios. Leveraging the comprehension and summarization prowess of the GPT4 model \cite{openai2024gpt4technicalreport}, we refine this data further. Subsequently, human evaluators curate and amend the data based on predefined criteria, yielding a collection of social scenarios featuring multiple participants and a spectrum of conflicts. The presence of heightened conflict is instrumental in rigorously testing the models' capabilities within intricate social contexts. This process culminates in the creation of a multi-user multi-turn dialogue dataset that is intrinsically linked to these scenarios.

In this study, we delve into the nature of evaluation methodologies for models in multi-user multi-turn social tasks and propose a novel metric HSII score. A recent study \cite{ren2024emergencesocialnormsgenerative} introduces a framework aimed at enhancing social norms. Furthermore, Mao et al. \cite{mao2024multiuserchatassistantmuca} emphasize the importance of selecting an interlocutor and crafting dialogue content in multi-turn conversations. This involves determining "who to converse with and what to convey," while also establishing an interaction pattern across multiple users through extended dialogues. Building on these insights, we have designed a four-tiered evaluation protocol: prompting the model to respond with certain format, enabling the model to select from a broader array of potential interlocutors within our curated multi-turn dialogue dataset, articulating transitions between turns, and ensuring the continuity and stability of the dialogue post-switch, all of whose results was compiled up to compute the final HSII score. In this track we carry out our experiments on LLMs and propose our discoveries.

Additionally, when talking about capabilities in certain social scenes, one may question how Chain of Thought (COT) \cite{Wei2022ChainOT} affect LLMs' performance, given the potential of COT to augment social interaction skills within social contexts, iterative reasoning loops \cite{xagent_projects}, and scholarly work suggesting that a well-crafted COT can address high-level challenges in a mathematical framework \cite{zhang2024diagramthought}. Hence in our benchmark we introduce another novel metric, COT complexity of LLMs under specific COT configurations. By assessing the minimum number of reasoning cycles a model must undergo under a thoughtfully crafted set of COT within the model's self-reflection to strike given accuracy threshold, we effectively benchmark the cognitive efficiency of various models.

We summarize our contributions in two main folds:
\begin{enumerate}
    \item We make investigation about more complex tasks in social life and propose a systematical formulation of multi-user multi-turn social tasks structure.
    \item We introduce the How Social Is It (HSII), a statistical metric for quantifying social capability in multi-user multi-turn complex task scenes based on theoretical derivation and sociological conclusions. We then present how the dataset construction and evaluation pipeline is extracted in detail.
    \item We introduce the COT complexity metric to measure how efficient LLMs are to do reasoning and reflection along given set of COTs to meet given standard. We develop a pratical pipeline for this evaluation.
\end{enumerate}

\section{Related Work}
\textbf{Social Relationship and Social Scene.} Crafting social scenarios and interactions involves several critical elements and stages. Social scenarios are intricately woven from components such as setting, participants, and behavioral norms\cite{goffman1959presentation}. The process of social interaction generally unfolds through stages including initiation, development, and termination\cite{duncan1972some}. Dialogue acts involve the selection of conversational targets and the management of transitions between different speakers\cite{clark1991grounding}. Navigating multi-turn dialogues with a single conversational target necessitates the application of adjacency pairs and the preservation of topical coherence\cite{clark1987collaborating}.

\textbf{Agent Task Stratification.} Within the academic community, significant progress has been made in the field of agent task stratification. This process involves the systematic breakdown of complex missions into smaller, more tractable subtasks, which is essential for the effective operation of Multi-Agent Systems (MAS).
For instance, Hierarchical Reinforcement Learning (HRL)\cite{Li2019HierarchicalRL} tackles the challenges posed by sparse rewards and complex environments by decomposing intricate tasks into simpler subtasks. Meta-Task Planning (MTP)\cite{Zhang2024MetaTaskPF}, a strategy for simplifying complex task planning in collaborative, LLM-based MAS, furthers this approach by decomposing tasks into a sequence of subordinate tasks, or meta-tasks, which are then translated into actionable steps.
AI Agents can be classified into a spectrum of levels, ranging from Level 0 (non-AI, basic tools) to Level 5 (highly advanced agents exhibiting personality and cooperative interaction)\cite{Huang2024LevelsOA}. Each ascending level integrates additional modules and functionalities, thereby augmenting the AI capabilities and utility of the agents. The SMART-LLM framework\cite{Kannan2023SMARTLLMSM} exemplifies this progression, applying LLMs to multi-robot task planning. It translates high-level directives into actionable multi-robot plans through a systematic process that includes task decomposition, coalition formation, and task allocation.

\textbf{LLM Intelligent Agent Application Evaluation and Benchmarking.} Evaluating large AI models necessitates a rigorous methodology to assess their performance, robustness, and reliability\cite{nist2022}. This evaluation is essential for guaranteeing that models adhere to benchmarks of safety, efficacy, and ethical standards prior to their operational deployment\cite{aiindex2024}. Prominent benchmarks for LLM assessment include ImageNet\cite{russakovsky2015imagenetlargescalevisual}, a seminal benchmark for computer vision models, GLUE\cite{wang2019gluemultitaskbenchmarkanalysis}, which evaluates natural language understanding, and ArenaBench\cite{Kastner_2022}, a comprehensive benchmark designed to assess AI systems across a spectrum of tasks and environments\cite{mlsys2020}. These benchmarks are instrumental in promoting transparency and propelling the evolution of AI technology.

\section{Preliminaries}
The primary focus lies in evaluating model efficacy in social tasks characterized by multi-turn dialogues within intricate scenarios, involving numerous conversational targets. In this study, we rigorously assess the performance of models on originally built multi-user multi-turn social task datasets HSII. These tasks necessitate a diverse array of capabilities and often entail intricate interactions among multiple participants, underscoring the importance of considering both factors. 

\textbf{Social Task capability Objectives Division.} \cite{zhou2024sotopiainteractiveevaluationsocial} designs a multi-dimensional framework with various objectives, including the following and so on:
\begin{itemize}
    \item \textbf{Goal Completion (GOAL)} This is the extent to which the agent achieves their goals.
    \item \textbf{Believability (BEL)} This focuses on the extent to which the agent's behavior is perceived as natural, realistic, and aligned with the agent's character profile, thus simulating believable proxies of human behavior.
    \item \textbf{Knowledge (KNO)} This captures the agent's capability to actively acquire new information.
    \item \textbf{Secret (SEC) [-10-0]} This measures the need for agents (humans) to keep their secretive information or intentions private.
    
\end{itemize}

\textbf{Framework of Multi-User Chat (MUC)}The multi-user framework architecture and information flow consist of three major modules: Sub-topics Generator, which generates the initial sub-topics; Dialog Analyzer, which extracts short-term and long-term features from chat history; Utterance Strategies Arbitrator, which determines the dialog acts corresponding to our design dimensions.

Overall, the Sub-topics Generation is executed once, and the Dialog Analyzer and Utterance Strategies Arbitrator are executed sequentially for every next utterance, which ensures latency-efficiency in front of higher message traffic and complex interactions from multiple users. Among them, the Dialog Analyzer consists of sub-modules containing Sub-topic Status Update, Utterance Feature Extractor, Accumulative Summary Update, Participant Feature Extractor. The Utterance Strategies Arbitrator includes modules like Direct Chatting, Initiative Summarization, Participation Encouragement, Sub-topic Transition, Conflict Resolution.\cite{mao2024multiuserchatassistantmuca}

\section{Framework}
Here, we delineate our social agent architecture, which is meticulously crafted.

\begin{wrapfigure}{r}{0.5\textwidth} 
\vspace{-0.2cm}
\centering
\includegraphics[width=0.45\textwidth]{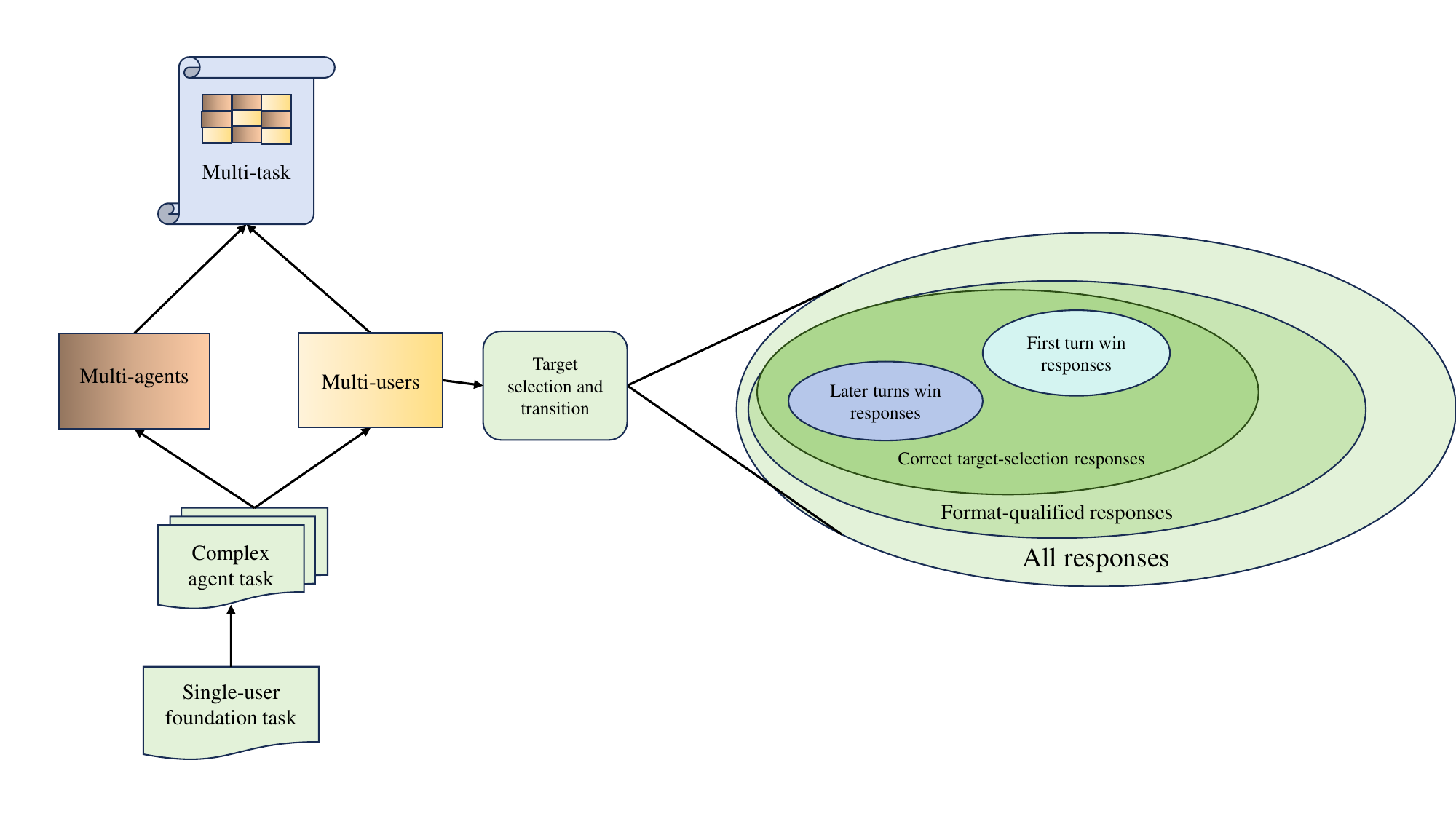} 
\caption{Main leveling of agent task and capability evaluation. On the left is different levels of social tasks including basic single-user tasks, multi-agents and multi-users tasks based on the first ones, and final multi-task ones on top.  We mainly put sight on multi-user tasks. Then on the right is our four-step evaluation framework for multi-user tasks.}
\label{dataset-construct}
\vspace{-0.4cm}
\end{wrapfigure}

Drawing inspiration from pivotal sociological theories that dissect social relationships \cite{abbott2020overview}, \cite{bondarenko2020social}, and \cite{tromp2022five}, we dissect social interactions into three critical facets:
\textbf{Object Transition} Identifying the subsequent conversational target during the object transition phase.\textbf{Transitional Utterance} Formulating and selecting dialogue content for the forthcoming interaction with the designated conversational target.\textbf{Post-Transition Multi-Turn and Multi-Level Dialogue} Engaging in multi-turn dialogues with the chosen conversational target post-transition, evaluating the aggregate effects and ultimate outcomes.

\subsection{Agent Task Leveling from the Social Agent Perspective}
Here we give brief construction for two agent task groups. More precise inside structure for both types are presented in appendix part.

\textbf{Single-User Foundational Agent Tasks} Within a predefined protocol, one agent addresses inquiries from only one single user and execute API call commands with given tools. The goal is to attain high precision in instruction following.The foundational approach entails a single-step agent call, while an alternative strategy is predicated on the Chain of Thought (CoT) \cite{Wei2022ChainOT}.

\textbf{Complex Agent Tasks} Expanding upon the single-user foundational task, we define composite tasks as those encompass two categories of agent enhancement that may interrelate and nest: multi-agents and multi-users. Multi-agents tasks stand for that multiple LLM agents collaborate to jointly accomplish a single task \cite{han2024llmmultiagentsystemschallenges}\cite{guo2024largelanguagemodelbased}\cite{wu2023autogenenablingnextgenllm}.Multi-users tasks ensemble that single LLM agent serves multiple users, necessitating the determination of the current conversational target, the dialogue content to facilitate target transition, and the optimization of outcomes for all targets post-transition. The final comprehensive multi-tasks mean single LLM agent performs diverse tasks for multiple users, exemplifying generalizability \cite{tan2023cappyoutperformingboostinglarge}\cite{chen2023tigerbotopenmultilingualmultitask}.

\subsection{Talking Target Transition in Multi-user Multi-turn Dialogue Systems}
In practical applications, dialogue systems often encounter scenarios where a single agent must engage with multiple distinct users, each requiring tailored responses. This dynamic unfolds across multiple dialogue turns, with target transitions facilitating the switch between interlocutors. For instance, in a school setting like a parent-teacher conference, various targets like parents, students, teachers, principals are involved. Certain information, such as a student's report card, is restricted to specific parties. A straightforward solution is to have a single intelligent assistant handle these diverse needs simultaneously, representing the temporal relationships through a unified target. However, challenges arise in complex social scenarios where the needs of multiple parties are interdependent. For example, an intelligent assistant may need to interact with the parents of high-achieving students and teachers to glean effective learning strategies first, which can then inform advice for parents of struggling students. In such cases, the assistant must assess the priority of responding to different targets, drawing on the social dynamics of real-world conversations \cite{choi2020tensocialdimensions} \cite{peralta2022opiniondynamicssocialnetworks}\cite{adams2022waysunderstandingsocialdynamics}, and generate utterances based on the unique information pertinent to the selected target. In our approach we focus on the evaluation of LLMs' performance in those complex scenes, separately in single-turn and with COT form, under different requirements.

\section{Methods}
In this section, we introduce the pipeline to build our HSII dataset and then we propose HSII evaluation framework for LLMs' capability in multi-user multi-turn social agent tasks based on HSII dataset.

\subsection{Construction of a Multi-user Multi-turn Dialogue Dataset from News Datasets}
\begin{figure}[t!]
    \centering
    \begin{subfigure}[b]{0.48\textwidth}
        \includegraphics[width=\textwidth]{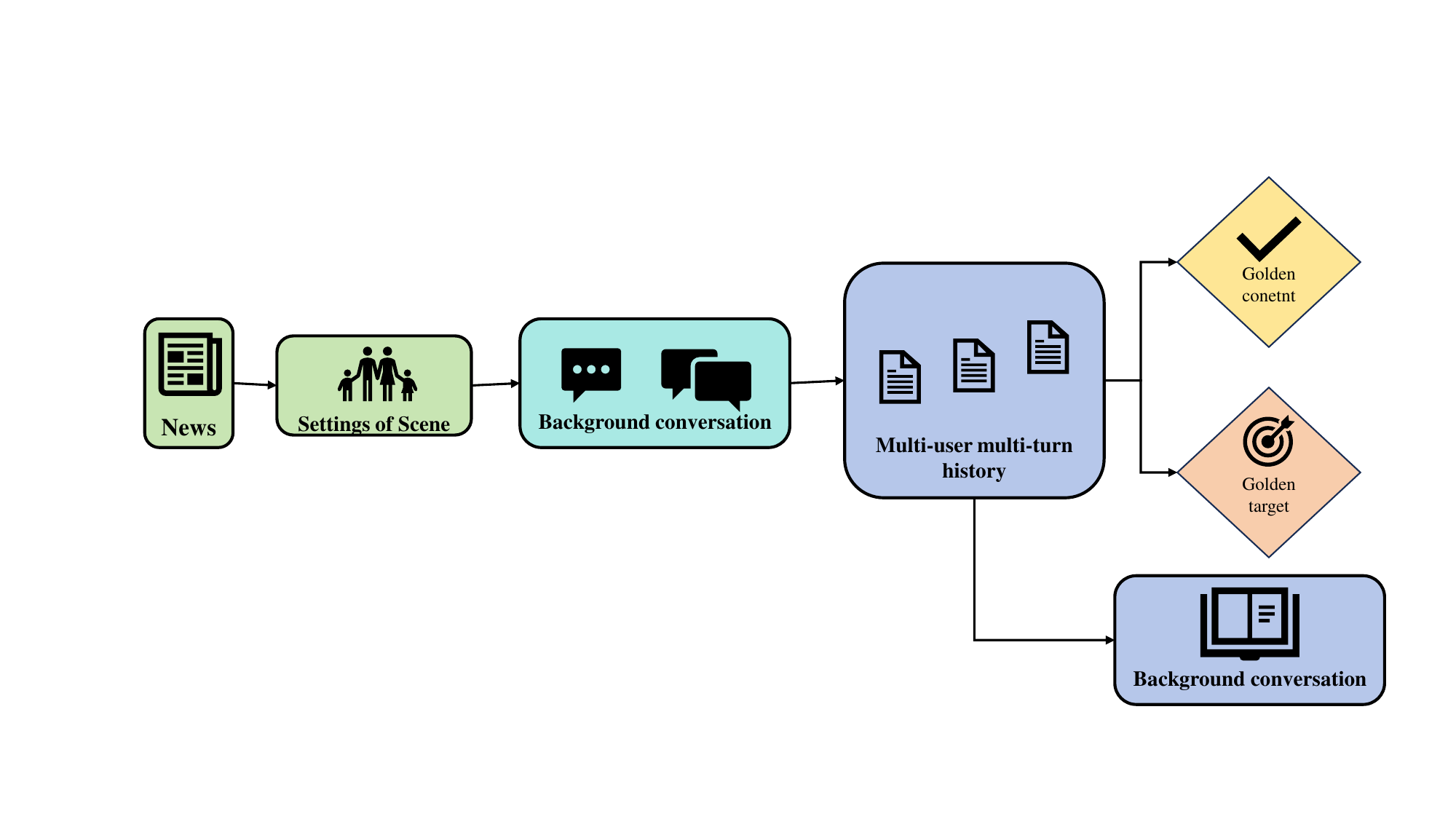}
        \caption{Evaluation dataset construction design.}
        \label{subfig1:build}
    \end{subfigure}
    \begin{subfigure}[b]{0.48\textwidth}
        \includegraphics[width=\textwidth]{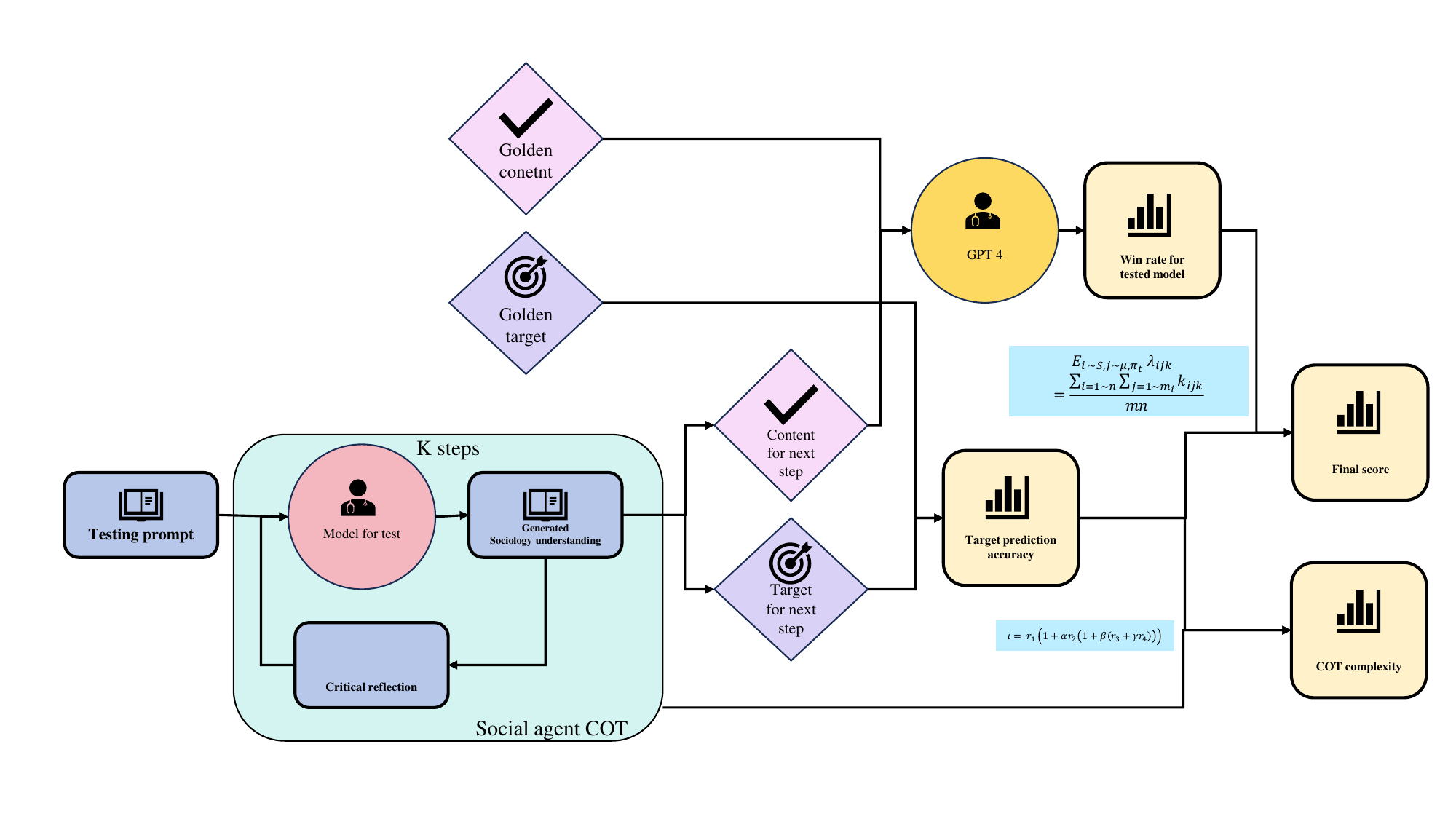}
        \caption{HSII evaluation framework pipeline.}
        \label{subfig1:evaluation}
    \end{subfigure}
    \caption{Evaluation dataset construction design and HSII evaluation framework pipeline.}
\end{figure}

Our approach begins with the strategic selection of one to two keywords to seed news searches, programmatically retrieving relevant news articles and documents\cite{leeb2024diversemultilingualnewsheadlines} \cite{gao2024generativenewsrecommendation}. These documents are then employed to craft thematic descriptions. Utilizing the thematic descriptions, we proceed to simulate dialogue data by meticulously extracting and organizing the pertinent thematic elements into scenario components with GPT4\cite{openai2024gpt4technicalreport}. In the final stage, these components are meticulously assembled to form comprehensive multi-user, multi-turn dialogues in HSII, by GPT4 and manual refining. To optimize the resource-intensive search process, we harness pre-processed offline news datasets as seeds for scenario generation.
In our pipeline to augment the dataset's authenticity, we curate news report excerpts from diverse sources that encapsulate real-world events and distill the key details and logical connections within the reports, transforming them into a structured background setup with multiple fields, including domain, brief Scene Description, main Scene Participants and Social Relationships, and Potential Conflicts Among Participants.The entire process is graphically represented \ref{subfig1:build} for clarity.

\subsection{Evalution Framework}
\subsubsection{Theoretical Settings}
\textbf{HSII overall score.} We introduce a novel metric, overall HSII score, to measure how social one tested LLM performs in four evaluation stages: verifying whether parsed response can suit requirements; selecting the next talking target in the social task scene; generating the first statement after switching; and engaging in sustained dialogue after switching, just as follows:
\begin{definition}
For each test case $s_i$ in the test dataset $S$ with size $n$, we input it to the model $\pi_t$ and get response $\mu_i$. Then we parse $\mu_i$ to required pattern. Then we count the rate of successful parsing $r_1=n_1/n$, then from parsed output dict we match target selection in this step $t_i$ with golden target $\textbf{t}_i$ and count $n_2 = \sum_{i=1}^n 1(t_i=\textbf{t}_i)$ to get successful target selection rate $r_2=n_2/n_1$. We input first utterance content $\omega_i$ and golden one $\Omega_i$ to GPT4 for judgment which wins, loses or equals, on reverse sides to avoid positional bias, getting win rate of the test model $r_3=n_3/n_2$. Finally we prompt testing model to chat for several turns. Similarly get long-run win rate $r_4=n_4/n_2$.
The final overall HSII score, noted as $\iota$, as 
\begin{equation}
    \iota = r_1(1+\alpha r_2(1+\beta (r_3+\gamma r_4)))
\end{equation}.
\end{definition}
$\alpha,\beta, \gamma$ in the equation should be experimental hyper-parameters for overall evaluation. Here we take weight $\alpha = 1.0$, $\beta=1.0$ and $\gamma=1.0$ as equal for each stage for fairness.
Apart from sociological background discussed earlier, approach of this metric ensures sufficient discrimination between similar test models. An overall analysis is provided in the appendix section. 

\textbf{COT complexity.} Previous work proposes COT boost LLMs' performance\cite{Wei2022ChainOT}. However it's simple that the COT methodology demands more computational resources than single-turn problem-solving. Also notably, longer COTs are computationally more intensive than their shorter counterparts. To quantitatively assess the efficiency of AI models, we introduce a nature metric: the social task complexity, as following. This metric evaluates a model's performance under specific COT designs when tackling certain questions.

\begin{definition}
Given a test dataset $S$ comprising $n$ test cases $s_i$, we construct a standardized COT set $\boldsymbol{\mu} = \{\mu_{i1}, \mu_{i2}, \ldots, \mu_{im_i}\}$ for each test case $s_i$, with a set size of $m_i$. This COT set serves as a guide for various models $\boldsymbol{\pi} = \{\pi_1, \pi_2, \ldots, \pi_K\}$ to deliberate and respond to queries. For a particular model $\pi_t$, when it produces an answer aligning with the golden standard for $s_i$ under a specific COT $\mu_{ij}$ after $k_{ij}$ iterations of reflection and guidance, the COT complexity $\lambda_{ijt}$ for this social task $s_i$ under $\mu_{ij}$ for $\pi_t$ is recorded as $k_{ijt}$. In scenarios where the problem's complexity surpasses the capabilities of the current COT-framework-model pair, the COT complexity $\lambda_{ijt}$ is regraded as infinite.The COT complexity for model $\pi_t$ across the dataset $S$ is then defined as the average COT complexity under all test queries and corresponding COTs, mathematically expressed as:
\begin{equation}
    E_{i \sim S, j \sim \boldsymbol{\mu}, \pi_t} \lambda_{ijt} = \frac{\sum_{i=1}^n\sum_{j=1}^{m_i}k_{ijt}}{mn}
\end{equation}.
\end{definition}

\subsubsection{Evaluation pipeline}
In our proposed evaluation framework HSII, the multi-user dialogue capabilities of a LLM agent are rigorously assessed through both objective and subjective measures. Main evaluation pipeline is displayed in \ref{subfig1:evaluation}.

The objective evaluation focuses on accuracy of target selection, which is quantified by calculating the proportion of correct next-target selections made by test LLM across all test cases. The subjective evaluation assesses the quality of the first-utterance and long-run statements generated by the model. Here we adopt the win rate metric introduced in ToolBench \cite{qin2023toolllmfacilitatinglargelanguage} to gauge the overall performance. There is one difference that we adopt both GPT4 and human-eval to get final win rate. Incorrect selections result in no score, as they lead to an invalid dialogue sequence by the LLM agent. But if in later phase the response is unfavorable, some score may still be awarded for correct selection. In appendix section we provide a rough theoretical foundation and sociology meaning for this approach.

\subsubsection{Human Evaluation Involved to Support HSII Metrics}
We involve human evaluation about HSII metrics in in two ways. As the first one, during dataset construction pipeline we employ human evaluation to do data cleaning, removing those golden responses that do not fit human value, to get the preference pairs exactly match human value judgment in certain complex social scenes. Further more, in evaluation pipeline where GPT is used to do adversarial evaluation, we also employ human evaluation by modifying and correcting the different win/lose judgment made by GPT that differ from human judgment. The goal of human evaluation in this part is to align value preference in built dataset the same as human value. 
On the reverse dimension, in evaluation experiments to draw HSII score for different models, we measure human performance in comparison to LLMs. Human responses in overall settings generated by a group of subjects get the highest, even close to full score, which says our metrics correspond to human analysis with credibility. We also discuss in which way the human generated response is the same as model-generated ones while in which ways human perform differently compared to LLMs in evaluation pipeline, in both examplified and in statistical analysis.

\section{Experiments}
\subsection{Evaluation Dataset Build}
Utilizing the methodology outlined we construct HSII dataset with two steps. We begin with generating scenarios that encompass target transitions. By employing top-k scenario sampling and ascertaining which scenarios accurately meet predefined criteria and mirror real-world complexities involving intricate social dynamics and conflicts, we refine the scenarios and craft representative multi-user multi-turn dialogue test cases. 

For more detail the characters involved in conversational scenario are greatly various for different scenes. For example, we mainly extract scenes from sports, politics, education, weather, research and business news set. In sports topic main characters contain athletes, refugees and audience. In education part they may be teachers, students, parents and so on while in business they may be modified to shoppers, customers, government workers and business manager. For better knowledge we provide additional statistics about the dataset. It contains $N_0=8305$ samples in total. On average in each sample there are $N_f=6.722$ unique characters and $N_c=7.801$ turns of conversation. But in fact the most significant difference of our dataset compared to existing multi-party chat datasets should be the explicit comparison of golden response and tested response of the "agent" character directly designed for LLM to play, in precise conversation position. Currently our dataset are under examination to avoid risk of leaking by sponsor company, but after the pipeline the complete version will be public. 

After systematically analyzing each sample's dialogue sequence and extract the preceding dialogue as contextual background, the assistant's response as golden responses. The backgrounds and golden responses' pairs are consolidated to constitute the final test sample set.

\subsection{Clustering analysis on our dataset}
\begin{wrapfigure}{r}{0.5\textwidth} 
\vspace{-0.6cm}
\centering
\includegraphics[width=0.45\textwidth]{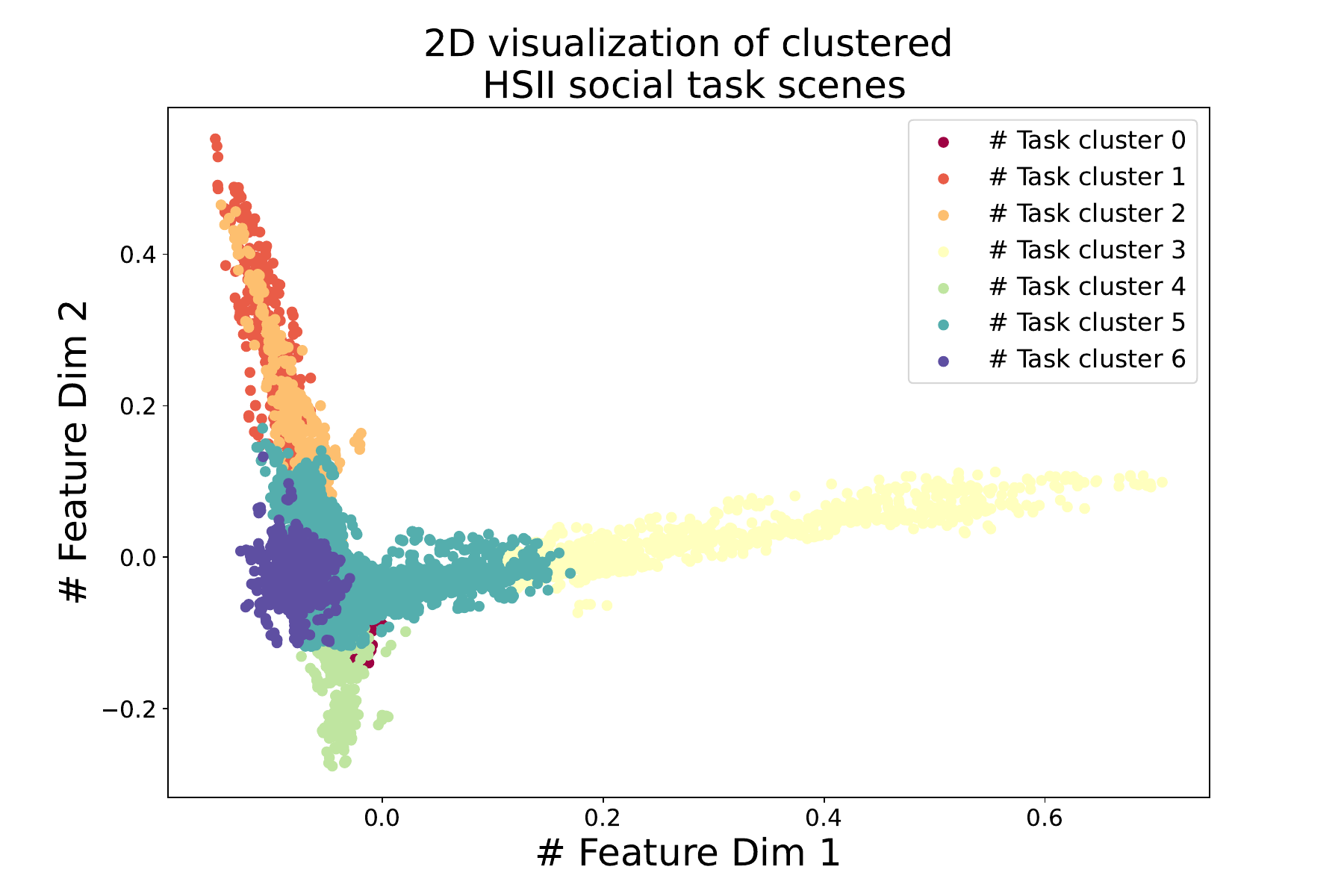} 
\caption{Clustering analysis of constructed dataset. Each color stands for one cluster of HSII dataset, mainly matching one field or paradox feature in social scenes.}
\label{cluster-analysis}
\vspace{-1cm}
\end{wrapfigure}
Furthermore, we conduct an analysis of HSII dataset to ascertain its breadth of coverage. Specifically, utilizing the BERT model \cite{devlin2019bertpretrainingdeepbidirectional}, we extract features from our test query cases. Then we apply the DBSCAN \cite{wang2019dbscanoptimalratesdensity} clustering method to them. After dimension reduction with LSH (Locality Sensitive Hashing) \cite{jafari2021surveylocalitysensitivehashing}, we visualize the cluster graph as shown in Figure \ref{cluster-analysis}.The clustering outcomes predominantly encompass seven dimensions, which exhibit both similarities and differences compared to types of original source news. 

\subsection{HSII evaluation for social capabilities}
\paragraph{Setup}
During evaluation, we first employ the multi-user multi-turn dialog as history, adhering to prompts detailed in appendix. The tested model (\(\pi\)) was assumed the role of an intelligent assistant to select its subsequent target. During this phase, we meticulously parse selected targets and dialogue utterances from \(\pi\)'s responses. We compare the chosen target name with golden standard, which covers all possible duplicated names for robustness, to get accuracy score. Consequently, we assess quality of dialogue utterance whose target selections are correct. After employing the traditional adversarial evaluation method by input \(\pi\)'s dialogue utterance and gold standard to GPT4 and human grader for scoring, we get the win rate of \(\pi\)'s responses. Finally, we combine accuracy of target selection with the win rate of responses and calculate overall HSII score.

\begin{table}
\centering
\begin{tblr}{
  row{even} = {r},
  row{3} = {r},
  row{5} = {r},
  row{7} = {r},
  column{1} = {c},
  cell{1}{2} = {c},
  vline{2} = {-}{},
  hline{1-2,7} = {-}{},
}
             & $r_1$ & $r_2$ & $r_3$ & $r_4$ & HSII \\
llama2-7b    & 0.472 & 0.510 & 0.26  & 0.27  & 0.600         \\
baichuan2-7b & 0.343 & \textbf{0.624} & 0.40  & 0.44  & 0.522         \\
qwen2.5-7b   & \textbf{0.677} & 0.266 & 0.47  & 0.52  & 0.855         \\
llama3-8b    & 0.554 & 0.565 & \textbf{0.55}  & \textbf{0.55}  & \textbf{0.898}         \\
mistral-7b   & 0.496 & 0.491 & 0.41  & 0.44  & 0.703         \\
GPT4         & 0.701 & 0.732 & 0.67  & 0.69  & 1.399         \\
human        & \textbf{0.996} & \textbf{0.804} & \textbf{0.72}  & \textbf{0.72}  & \textbf{2.149}         
\end{tblr}
\caption{Evaluation result of major LLMs on our bench. $r_1$, $r_2$, $r_3$, $r_4$ and $HSII$ specifically account for format passing rate, absolute target selection pass rate, relative score(win rate), relative score(win rate) in the long ($\epsilon=7$) run and overall HSII score. The best performance in model groups with size relative size is \textbf{bolded}.}
\label{table1}
\end{table}

For limitation of computility currently we employ models with relative size including Llama2-7b\cite{touvron2023llama2openfoundation}, baichuan2-7b\cite{yang2023baichuan2openlargescale}, qwen2.5-7b\cite{hui2024qwen25codertechnicalreport}, llama3-8b\cite{dubey2024llama3herdmodels}, mistral-7b\cite{jiang2023mistral7b}. We also benchmark online LLM GPT4\cite{brown2020languagemodelsfewshotlearners} and real humans, who are not involved in dataset construction, by quantifying average score of their responses in comparison, as depicted in \ref{table1}.

\paragraph{Result}
We analyze models' interactions performance on HSII to assess their social capabilities in multi-user multi-turn social tasks. Figure \ref{table1} displays average rate tested models adhere to required format, absolute target selection pass rate, the relative score orwin rate in comparison to golden answer provided by GPT-3.5 both in first utterance and longer range, and the overall HSII score. In general, GPT-4 consistently outperforms all other LLMs across all four phases ($\sim$0.03, $\sim$0.11, $\sim$0.12, $0.14$). Among models of relative size, Llama3-8b albeit with a lower format pass rate ($\sim$0.12) than Qwen2.5-7b and a lower target selection accuracy ($\sim$0.06) than Baichuan2-7b. However, Llama3-8b scores higher ($\sim$0.08, $\sim$0.03) than the latter two models in both win rate score. This underscores the significance to evaluate models' social abilities across all our multiple dimensions. Following these top performers are Mistral-7b and Llama2-7b. For supplement below we further present more discoveries.

\textbf{Human responses still lead the way.} In our evaluation benchmark, human responses maintain a clear advantage over LLMs, including GPT4. This suggests that there may be a persistent discrepancy in action patterns between humans and current LLMs in complex social scenes. Results reveal humans often exhibit more straightforward behavior with changes in talking target. For example, in scenario to purchase food within a budget, humans promptly approach the salesperson to inquire about prices, which is typically preferred in real-world interactions, whereas LLMs tend to redundantly seek clarification on details specified in previous instructions. We may say sometimes an overemphasis on 100\% accurate quoting and logical reasoning do not exactly align with complex practices in reality.

\textbf{Models try employing tricks to bypass explicit conflicts.} We observed that certain models, especially GPT4, occasionally produce peculiar responses, attempting to circumvent conflicts in social scenarios. For instance, when prompted to relay unfavorable information to a student's family, the LLM only provides a brief overview of the situation before quickly shifting focus to more positive imaginations, rather than directly engaging with the parents about the details.

\textbf{LLMs exhibit more challenges in first utterance than in the long run.} Our result indicates that the model's performance in first utterance subsequent to target transition is consistently inferior to that in longer run with average gap of $0.025$ across all LLMs. This observation underscores the models' difficulty in swiftly adapting to new background and context following transition in talking target. A possible explanation may lie in after engaging with a target over several rounds, the model activates target-specific knowledge within the social context, facilitating appropriate responses, while in first utterance post-transition the model grapples with the abrupt transition, with knowledge base still rooted in last target. This suggests preemptively summarizing former conversation after transition, as proposed in \cite{liu2024learningsummarizelargelanguage}\cite{wan2024fusechatknowledgefusionchat}, may mitigate this issue.

\subsection{Will llms do better with more prompting?}
\paragraph{Setup}
\textbf{Major Changes with COT Added} We implement the decomposition of complex instructions through specific COT structures. This approach provides the model with more precise and specific prompts per sub-task, directing its focus towards key points and simplifying comprehension.
The major change take place extra analysis of given scene(task) for LLM. In fact as presented in Figure 4 we designed a COT cycle involving clarifying current psychological states of related targets in the scene as first stage. Then the LLM tested analyze demands and motivations of those targets to infer what they may think; judge whether the demands of those targets can be satisfied together while which will be in conflict. With those conflicts in mind the models tested are ordered to think about which conflict or demand should be settled first and then the LLM is guided to really do the job, generating talking target next turn and talking content. This chain fulfills a complete design-making cycle. However in the process agent may encounter ignored demands, conflicts and relationships. So we induct reflection method. After the LLM give out final chosen target and generated words we instruct it to reflect whether the target choice and words may have drawbacks in alignment dimensions "willing to help", "professional", "harmless" and "empathetic". If yes, put this understanding into the first part of COT cycle as extra psychological states and start a new cycle. In COT score evaluation patch we do this reflection pipeline by turns, until in one turn chosen target is modified to the one in golden response.

An example is provided below.

\begin{figure*}[!th]
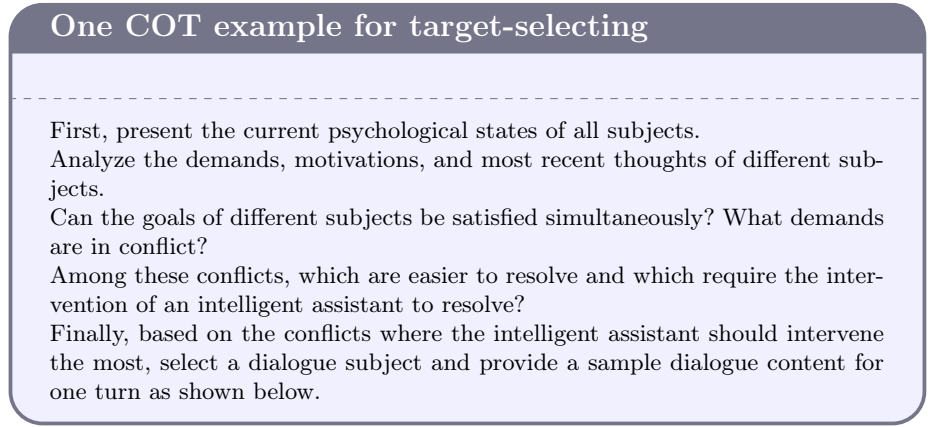

\begin{tcolorbox}[
    colback=blue!6!white, 
    colframe=blue!8!gray, 
    title=\textbf{One COT example for target-selecting}, 
    fonttitle=\bfseries\large, 
    arc=4mm, 
]
\tcblower
First, present the current psychological states of all subjects.\\
Analyze the demands, motivations, and most recent thoughts of different subjects.\\
Can the goals of different subjects be satisfied simultaneously? What demands are in conflict?\\
Among these conflicts, which are easier to resolve and which require the intervention of an intelligent assistant to resolve?\\
Finally, based on the conflicts where the intelligent assistant should intervene the most, select a dialogue subject and provide a sample dialogue content for one turn as shown below.
\label{COT example}
\end{tcolorbox}
\caption{One example in our COT set.}
\end{figure*}

\begin{table}
\centering
\begin{tblr}{
  row{even} = {r},
  row{3} = {r},
  row{5} = {r},
  row{7} = {r},
  column{1} = {c},
  cell{1}{3} = {c},
  vline{2} = {-}{},
  hline{1-2,8} = {-}{},
}
             & {Success rate \\without COT} & {COT steps \\needed for \\success rate 0.70} & {Success rate \\with COT} & COT complexity \\
llama2-7b    & 0.510                        & 22.6                                         & 0.552                     & 38.4           \\
baichuan2-7b & 0.624                        & 20.8                                         & 0.650                     & 33.1           \\
qwen2.5-7b   & 0.266                        & 18.4                                         & 0.441                     & 35.8           \\
llama3-8b    & 0.565                        & 14.9                                         & 0.619                     & 29.5           \\
mistral-7b   & 0.491                        & 17.9                                         & 0.539                     & 34.4           \\
GPT-4        & 0.732                        & 10.1                                         & 0.787                     & 27.6           
\end{tblr}
\caption{Evaluation of COT complexity.}
\label{COT-result}
\end{table}

To quantitatively evaluate various models and mitigate impact of exceptionally challenging problems involving intricate social dynamics that even humans might struggle to navigate optimally on the final outcome, here we impose an upper limit of $N_{\infty} = 128$ on the number of reasoning and reflection rounds. This cap ensures the results not disproportionately swayed by these extreme scenarios. Full results are displayed in Table \ref{COT-result}.

\paragraph{Result}
Table \ref{COT-result} reveal that incorporating a 6-step COT reasoning into our experimental framework leads to a plausible improvement in model performance on HSII with an average lead of $\Delta=0.067$. Among these, the highest improvement observed is $\Delta_{max}=0.175$ from qwen2.5-7b. This approach has narrowed the gap with human response, although responses from LLMs still fall short of ones from real human. More than that, with COT complexity measurement we uncover more features.

\textbf{Simple COT can not cover all} Despite the utility of COTs, we have noticed some instances where object selection tasks exhibit persistent inaccuracies. Specifically, continuous COTs and reflections fail to achieve further optimization in those  target selection cases, leading to an escalated complexity for the models. To elucidate this phenomenon, we conduct an ablation study to assess the rounds of COT and reflection required for LLMs to reach an average selection accuracy threshold of 0.70, denoted as partial-COT. Our findings in table \ref{COT-result} indicate that the incremental rounds necessary for performance enhancement with same scale beyond greater threshold exceed those beyond smaller threshold because of those hard-to-solve cases, displaying an increasing challenge or bottleneck. 

\textbf{Greater Gains for Laggards.} Our observations reveal a notable variability in the extent of improvement across models under COT. Models initially performing suboptimally exhibit a more pronounced improvement post-COT compared to their counterparts with higher initial accuracy. For instance, the qwen2.5-7b model with an initial score of $0.266$ demonstrated a significant improvement of $0.175$ after COT implement, whereas one of the top-performing model baichuan2-7b only experienced a marginal enhancement of $0.042$. This disparity aligns with the optimization bottlenecks discussed above, where high-performing models encounter greater difficulty in surmounting the challenges posed by more complex queries even with COT.

\textbf{COT Complexity as a Discriminative Metric.} A comparative horizontal analysis reveals that the variance among different models is usually more pronounced when measured by COT complexity than by single-turn accuracy metric. This suggests that COT complexity may provide a new expressive evaluation metric, particularly in tasks involving target selection and complex decision-making pipelines.

\section{Conclusion}
In this research, we focus on evaluating the social communication capabilities of Large Language Models (LLMs) within multi-user, multi-turn real-world social contexts. To enhance our assessment of model adaptation to social scenarios and to potentially facilitate the integration of LLMs into real-life applications, we develop a novel framework HSII. This framework is grounded in traditional sociological theory and designed for overall social scenes. It complements the basic single-turn social evaluations with the complex scenarios. We harness the untapped potential of news source data to create the first multi-user multi-turn dataset that extensively covers real-life dialogue scenarios characterized by complexity and conflicts among various personas. Furthermore, we introduce a new statistical metrics, termed How Social Is It (HSII) overall score, to quantify LLMs' capability in navigating the challenging social scenes. This metric is derived from the discrimination bound of grading models at different stages. Then our focus also extends to the approach of COT to enhance model performance, an methodology that has been neglected in some previous benchmarks. To this end, we define a second novel metric, COT complexity, to measure the efficiency of LLMs when prompted and reflecting on certain social scenarios under a set of COTs. Based on the construction above, We detail the construction pipeline of our dataset and elucidate the workings of the entire evaluation process. Subsequently, we conduct evaluations on our benchmark using several representative LLMs and compare their performance with human beings, yielding novel and fresh results from these experiments. Looking forward, a compelling direction for future work is to expand the scale of our dataset and to test LLMs with a more diverse range. Additionally, probing the current capabilities of LLMs in social contexts presents a promising path for gaining insights into how LLMs perceive different characters, the roles they should assume in society, and how these roles might evolve.

\end{document}